\documentclass{article}

\usepackage{microtype}
\usepackage{graphicx}
\usepackage{subcaption}
\usepackage{booktabs} 

\usepackage{hyperref}

\usepackage[preprint]{icml2026}

\usepackage{amsmath}
\usepackage{amssymb}
\usepackage{mathtools}
\usepackage{amsthm}

\usepackage{booktabs}
\usepackage{multirow}
\usepackage{multicol}
\usepackage{enumitem}
\usepackage{xcolor}
\usepackage{soul}

\usepackage[capitalize,noabbrev]{cleveref}

\theoremstyle{plain}
\newtheorem{theorem}{Theorem}[section]
\newtheorem{proposition}[theorem]{Proposition}

\theoremstyle{definition}
\newtheorem{definition}[theorem]{Definition}

\theoremstyle{remark}

\usepackage[textsize=tiny]{todonotes}

\newcommand{\GP}{\textsc{Gold Panning}}
\icmltitlerunning{\GP: Strategic Context Shuffling for Needle-in-Haystack Reasoning}

\begin{document}

\twocolumn[
  \icmltitle{\GP: Strategic Context Shuffling for Needle-in-Haystack Reasoning}



  \icmlsetsymbol{equal}{*}

  \begin{icmlauthorlist}
    \icmlauthor{Adam Byerly}{jhu}
    \icmlauthor{Daniel Khashabi}{jhu}
  \end{icmlauthorlist}

  \icmlaffiliation{jhu}{Johns Hopkins University}

  \icmlcorrespondingauthor{Adam Byerly}{abyerly2@jhu.edu}

  \icmlkeywords{Machine Learning, ICML}

  \vskip 0.3in
]



\printAffiliationsAndNotice{}  

\begin{abstract}
Large language models (LLMs) exhibit pronounced position bias in long-context needle-in-haystack problems, systematically prioritizing the location of information over its relevance.
While current mitigations rely on \emph{white-box} access, this is effectively impossible for many state-of-the-art models.
We introduce \GP{}, a \emph{black-box} Bayesian framework that performs inference-time active search over long contexts by (i) reordering documents to concentrate high-belief items in highly diagnostic positions (signal anchoring) and (ii) updating beliefs over document relevance from model outputs.
Unlike conventional active learning, which prioritizes uncertainty reduction, \GP{} leverages anchoring---once flagged, keep it in sight---to preserve weak cues.
We implement this using iterative assignment derived from the model's diagnosticity profile, which provably identifies a target among $N$ documents in $O(\log N)$ rounds, ensuring scalability to many-document settings.
On needle-in-a-haystack retrieval and long-context QA, \GP{} matches Permutation Self-Consistency's target identification with $30\text{--}65\%$ fewer queries and remains effective under calibration mismatch, suggesting coarse positional ordering drives performance gains.
These results demonstrate that inherent model biases need not be failures, but can be used as tools for control. 
\end{abstract}

\section{Introduction}
\label{sec:Introduction}

\begin{figure*}[ht]
    \centering
    \scalebox{1}[0.943]{\includegraphics[width=0.95\linewidth,trim=0.0cm 1.0cm 0cm 0.7cm]{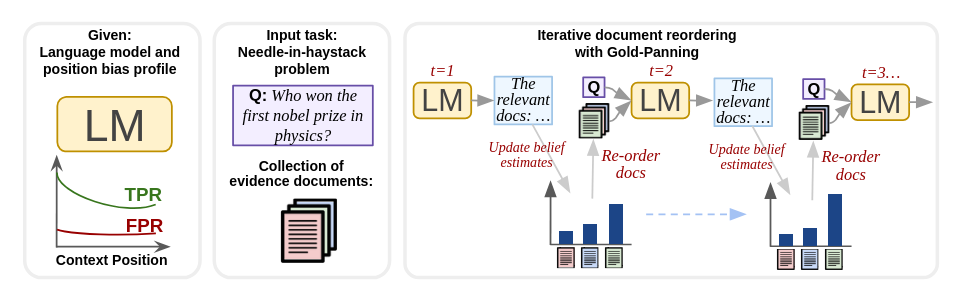}}
    \caption{Overview of \GP{}.
\textbf{Given:} a language model with a measured positional diagnosticity profile (TPR/FPR by position; \S\ref{subsec:diagnosticity}).
\textbf{Task:} identify relevant information from a document collection (\S\ref{subsec:Setup}).
\textbf{Method:} iteratively query the model, update Bayesian beliefs based on observed responses (\S\ref{subsec:beliefs}), and reorder documents to place high-belief items in high-diagnosticity positions (\S\ref{subsec:scoring}).
This accelerates belief separation, identifying relevant content with fewer queries than stateless ensembling.
}
    \label{fig:teaser}
    \vspace{-1em}
\end{figure*}

Many reasoning tasks require identifying relevant signals within large contexts, including scientific synthesis, legal analysis, and agentic fact retrieval.
LLMs have emerged as practical tools for these tasks, enabling knowledge synthesis in retrieval-augmented generation (RAG)~\citep{lewis2020retrieval, gao2023retrieval} and agentic systems~\citep{wang2023voyager, yao2023react}, but a critical bottleneck undermining reliability is \emph{position bias}, where LLMs often prioritize information based on \emph{where} it appears in context rather than its intrinsic relevance~\citep{wang2023primacy, zheng2024large, liu2024lost}.
Even state-of-the-art retrieval systems~\citep{ke2025large, zhang2025survey} cannot guarantee optimal placement.
When critical information falls into low-attention zones, reasoning failures increase substantially.

Current approaches to mitigate this bias fall into two categories.
\textbf{White-box mitigations} modify model architectures to reduce bias~\citep{peysakhovich2023attention, hsieh2024found, wang2024eliminating, zhang2024found}, but require model access or extensive fine-tuning, rendering them impractical for proprietary API models.
\textbf{Inference-time ensembling}, such as Permutation Self-Consistency (PSC; ~\citet{tang2024found}), offers a black-box alternative by randomly shuffling documents across multiple queries and aggregating results via majority vote.
While PSC averages over random placements, it assumes independent error, an assumption violated in long-context settings where position bias induces strongly correlated failures~\citep{byerly2025selfconsistency}.
Furthermore, PSC is inherently \emph{stateless}, as each shuffle is drawn \emph{independently}, discarding information from previous queries.
Finally, PSC is bias-agnostic, treating all positions as exchangeable rather than exploiting known positional reliability differences.

Rather than treating position bias as noise to average away, we propose to \emph{model it explicitly} and \emph{learn from observations}.
We introduce \GP, a black-box, inference-time framework that maintains Bayesian beliefs over document relevance and iteratively reorders inputs to align promising documents with positions of high discriminative power.
We observe that position bias has a consistent structure (Fig.~\ref{fig:teaser}, left pane), with some positions (e.g., the beginning and end of context) reliably surfacing relevant content, while others (the ``lost in the middle'' zone) tend to obscure it.
We characterize this structure by estimating a position's \emph{diagnosticity} (\S\ref{subsec:diagnosticity}), a measure of how reliably a position distinguishes relevant from irrelevant documents.
We perform a lightweight, per-model calibration (\S\ref{subsec:Calibration}, \S\ref{app:calibration}) using synthetic needle-in-haystack instances to estimate this positional reliability profile.
Empirically, we find this profile transfers across multi-document question answering (MDQA) tasks when the protocol is fixed (\S\ref{subsec:Transferability}).

Equipped with this profile, we re-imagine the context window not as a passive input buffer, but as an array of noisy detectors with known reliability.
By adopting a Bayesian lens (Fig.~\ref{fig:teaser}, right pane), we maintain a running probabilistic belief about each document's relevance, treating every LLM output as a noisy measurement rather than a final answer.
This transforms the problem from stateless ensembling into \emph{sequential Bayesian inference}, where each query extracts partial evidence that actively shapes the next input configuration, allowing the system to ``pan for gold.''

As document retrieval is fundamentally a \emph{search} problem, we do not need perfect belief calibration for all documents---we only need to efficiently identify the relevant ones. 
Consequently, reducing uncertainty about irrelevant documents yields only marginal gain towards this goal.
In long-context LLM reasoning, the search objective dominates: placing high-belief documents in high-diagnosticity positions consistently outperforms probing uncertain ones.

The underlying mechanism behind this is \emph{signal anchoring}.
Beliefs in long-context reasoning are inherently fragile; anchoring increases the probability of receiving high-diagnostic evidence early, which accelerates belief separation and reduces the number of rounds to confidently select the needles.
This principle, that \textbf{once identified, a needle must be kept in view,} maximizes recall with fewer queries.

\GP{} operationalizes these insights via greedy belief-diagnosticity matching at each round.
We evaluate on MDQA, a controlled setting where position bias is well-documented and retrieval success is directly measurable.
Across configurations, \textsc{GP-Belief} outperforms \textsc{GP-Entropy}, confirming that signal anchoring dominates uncertainty reduction.
Compared to PSC, \GP{} identifies relevant documents using \textbf{30--65\% fewer queries} while maintaining $F_1$ across model families and scales.

\textbf{Contributions.}
(1) \textbf{Framework:} We formalize position bias as a \emph{diagnostic signal} and introduce \GP{}, a black-box inference-time method for Bayesian active search over long contexts via calibrated position diagnosticity.
(2) \textbf{Theory:} We prove that a greedy belief-based assignment identifies targets in $O(\log N)$ LLM queries, reducing the cost of inference.
(3) \textbf{Finding:} \textsc{GP-Belief} outperforms \textsc{GP-Entropy} across all configurations, confirming that document retrieval is fundamentally a search problem, not a learning problem.
(4) \textbf{Empirical:} On MDQA benchmarks, \GP{} matches PSC's $F_1$ with $30\text{--}65\%$ fewer queries and remains effective under calibration mismatch, confirming that coarse positional ordering suffices.


\section{The \GP{} Framework}
\label{sec:Framework}

Our goal is to exploit position bias by repeatedly reordering documents across rounds to reliably surface relevant content with as few LLM queries as possible.
We first present the algorithm (\S\ref{subsec:algorithm}), then formalize each component: documents and positions (\S\ref{subsec:documents}), belief dynamics (\S\ref{subsec:beliefs}), positional diagnosticity (\S\ref{subsec:diagnosticity}), and scoring objective (\S\ref{subsec:scoring}).

\subsection{Overview and Algorithm}
\label{subsec:algorithm}

\GP{} operates in rounds.
Each round, we (1) score documents based on current beliefs,
(2) construct an assignment matching high-scoring documents to high-diagnosticity positions,
(3) query the LLM with this arrangement, and
(4) update beliefs based on which documents were cited.
Algorithm~\ref{alg:gp} summarizes this procedure:
\textbf{Lines 3-4} compute positional diagnosticity and establish a fixed ordering of positions by discriminative power (\S\ref{subsec:diagnosticity});
\textbf{Line 5} initializes uniform prior beliefs;
the main loop (\textbf{Lines 6-13}) iterates: \textbf{Line 7} scores documents according to one of two objectives (\S\ref{subsec:scoring}), \textbf{Lines 8-9} construct the greedy assignment matching document and position orderings, \textbf{Lines 10-11} format and query the LLM, and \textbf{Line 12} performs the Bayesian belief updates (\S\ref{subsec:beliefs}).

\subsection{Documents and Positions}
\label{subsec:documents}

We consider $N$ documents indexed by $i\!\in\!\{1, \ldots, N\}$, each with an unknown binary relevance $Z_i\!\in\!\{0, 1\}$, where we wish to identify the subset of documents with $Z_i\!=\!1$ (relevant) using a limited number of LLM calls.
For each query, we also have $N$ context positions indexed by $j\!\in\!\{1, \ldots, N\}$, which we treat as binary \emph{detectors} with true/false positive rates $\mathrm{TPR}_j, \mathrm{FPR}_j$.
Letting $O_{ij}\!\in\!\{0, 1\}$ denote if the LLM cites document $i$ at position $j$, we have:
\begin{equation}
\label{eq:observation_model}
    \Pr(O_{ij}\!=\!1 \mid Z_i\!=\!z) = z\!\cdot\!\mathrm{TPR}_j + (1\!-\!z)\!\cdot\!\mathrm{FPR}_j,
\end{equation}
with $\Pr(O_{ij}\!=\!0\!\mid\!\cdot)\!=\!1\!-\!\Pr(O_{ij}\!=\!1\!\mid\!\cdot)$.
To ensure structured outputs, we enforce a JSON citation schema via constrained decoding when available, otherwise via format-constrained prompting and validation.
The parameters $(\mathrm{TPR}_j, \mathrm{FPR}_j)$ are calibrated once (\S\ref{app:calibration}) and reused across tasks.
In Algorithm~\ref{alg:gp}, input documents $D$ (\textbf{Line 1}) are arranged into a prompt $P_t$ each round (\textbf{Line 10}).

\subsection{Actions, Observations, and Beliefs}
\label{subsec:beliefs}

\textbf{Observations.}
At each round $t$, we choose a \emph{permutation} $\sigma_t \in S_N$ (the symmetric group on $\{1, \ldots, N\}$) specifying which document is placed in which position; we interpret $\sigma_t(i)=j$ as ``document $i$ is placed in position $j$.'' 
After querying the LLM, we observe $O_{t,i} \in \{0, 1\}$ indicating whether document $i$ was cited at round $t$.
Let $j = \sigma_t(i)$ denote the position assigned to $i$.
The observation model (Eq.~\eqref{eq:observation_model}) applies with $j = \sigma_t(i)$.

Concretely, the model returns a set of cited document IDs $C_t$.
Given $\sigma_t$, we observe one outcome per document, $O_{t,i} = \mathbb{I}[i \in C_t]$, for $i \in \{1,\ldots,N\}$.

\textbf{Belief updates.}
We maintain posterior beliefs $\mathbf{b}_t = (b_{t,1}, \ldots, b_{t,N})$, where $b_{t,i} = \Pr(Z_i = 1 \mid \mathcal{F}_t)$ is our belief that document $i$ is relevant given all observations up to round $t$.
Starting from a uniform prior, $b_{0,i} = 0.5$, we update these beliefs via Bayes' theorem after each round using the observation model in Eq.~\eqref{eq:observation_model}:
\begin{equation}
\label{eq:bayesian_update}
    b_{t,i}
    =
    \frac{
        b_{t-1,i} \mathcal{P}_1
    }{
        b_{t-1,i} \mathcal{P}_1
        +
        (1 - b_{t-1,i}) \mathcal{P}_0
    },
\end{equation}
where $\mathcal{P}_z=\Pr(O_{t,i}\mid Z_i=z,\sigma_t(i)=j)$ (Eq.~\eqref{eq:observation_model}) is the Bernoulli likelihood under label $z$ for $j=\sigma_t(i)$.
When upstream retrieval scores are available, these could inform a non-uniform prior; we adopt the uninformative case for generality and to isolate the contribution of positional inference.
In Algorithm~\ref{alg:gp}, beliefs are initialized uniformly (\textbf{Line 5}) and updated after each query via Eq.~\eqref{eq:bayesian_update} (\textbf{Line 12}).

We define the log-odds
$
    \lambda_{t,i} = \log \frac{b_{t,i}}{1 - b_{t,i}},
$
which admits an additive Bayesian update (Eq.~\eqref{eq:bayesian_update}) of the form
$
    \lambda_{t,i} = \lambda_{t-1,i} + \ell_{\sigma_t(i)}(O_{t,i}),
$
where $\sigma_t(i)$ is the position assigned to document $i$ at round $t$, and $\ell_j(o)$ is the position-dependent log-likelihood ratio $
    \ell_j(o) = \log \frac{\Pr(O=o \mid Z=1, \text{pos}=j)}{\Pr(O=o \mid Z=0, \text{pos}=j)}.$
Thus, the log-odds evolve as a random walk with increments determined by the assigned position.

We adopt a conditional-independence approximation where citations are treated as independent given relevance and position.
While LLM decoding is autoregressive, we interpret $(\mathrm{TPR}_j,\mathrm{FPR}_j)$ as marginal rates under a fixed prompting protocol.
Empirical results (\S\ref{sec:Experiments}) support this approximation.

\begin{algorithm}[tb]
\caption{\GP}
\label{alg:gp}
{\footnotesize
\begin{algorithmic}[1] 
    \STATE {\bfseries Input:} Query $Q$, Documents $\mathcal{D}=\{d_1,\ldots,d_N\}$, Detector Profile $(\mathrm{TPR}_j, \mathrm{FPR}_j)$, Rounds $T$
    \STATE {\bfseries Output:} Final beliefs $\mathbf{b}_T$
    \STATE Compute $d_{\text{diag}}(j) \gets |\mathrm{TPR}_j - \mathrm{FPR}_j|$
    \STATE $\pi_{\text{pos}} \gets \text{argsort}(\mathbf{d}_{\text{diag}}, \text{descending})$
    \STATE Initialize $b_{0,i} \gets 0.5$
    \FOR{$t = 1$ {\bfseries to} $T$}
        \STATE \colorbox{yellow!30}{$s_i \gets \textsc{Score}(b_{t-1,i})$} \label{line:scoring}
        \STATE $\pi_{\text{doc}} \gets \text{argsort}(\mathbf{s}, \text{descending})$
        \STATE $\sigma_t \gets \text{zip}(\pi_{\text{doc}}, \pi_{\text{pos}})$
        \STATE $P_t \gets \text{Format}(Q, \text{Permute}(\mathcal{D}, \sigma_t))$
        \STATE $C_t \gets \text{QueryLLM}(P_t)$
        \STATE Update $\mathbf{b}_t$ via Eq.~\eqref{eq:bayesian_update} using citations $C_t$
    \ENDFOR
    \STATE {\bfseries return} $\mathbf{b}_T$
\end{algorithmic}
}
\end{algorithm}

\subsection{Diagnosticity}
\label{subsec:diagnosticity}

We quantify how informative a position $j$ is via Youden's index~\citep{peirce1884numerical, youden1950index}, a measure for assessing the discriminative power of diagnostic procedures.
Youden's $J$-statistic is defined as $J_j = \mathrm{TPR}_j - \mathrm{FPR}_j.$
Under this measure, a position is maximally informative when $J_j = 1$, uninformative when $J_j = 0$, and maximally \emph{anti}-informative when $J_j = -1$.
Crucially, when $J_j < 0$, a citation from position $j$ is more likely under irrelevance than relevance, and therefore provides strong \emph{negative} evidence.
Because we primarily need to decide \emph{which positions yield the most information}, we rank positions by \emph{magnitude}, $d_{\text{diag}}(j) = |J_j|$, while retaining the sign of $J_j$ through the likelihoods $(\mathrm{TPR}_j, \mathrm{FPR}_j)$ in our Bayesian update (Eq.~\eqref{eq:bayesian_update}).

For simplicity, we present the balanced case with $N$ documents and $N$ positions.
Asymmetric cases where the number of retrieved documents differs from the number of usable positions can be reduced to this setting by adding \emph{dummy} positions or items; we describe these details in Appendix~\ref{app:asymmetric}.

\subsection{Scoring Objectives}
\label{subsec:scoring}

\GP{} is a greedy policy that uses calibrated positional diagnosticity and evolving document beliefs to construct assignments.
    
\noindent \textbf{Document scoring.} At each round, we assign a scalar score $s_{t,i}$ to each document $i$ based on its current belief $b_{t-1,i}$, sort documents in descending order, and match the highest-scoring document to the most diagnostic position, second-highest to second-most diagnostic, and so on.
The choice of scoring function reflects two natural objectives corresponding to active learning~\citep{settles2009active, settles2011theories} and active search~\citep{garnett2012bayesian}. 
The first, \textbf{Uncertainty Reduction,} prioritizes exploring ambiguous documents to maximize information gain, assigning the most diagnostic positions to \emph{high-entropy} documents.
This follows standard intuition from active learning: probe the most uncertain cases to learn fastest.
The second, \textbf{Target Identification,} prioritizes confirming likely positives to maximize recall, assigning diagnostic positions to \emph{high-belief} documents.
We implement these objectives via two distinct scoring variants:

\noindent \textbf{\textsc{GP-Entropy} (uncertainty reduction).}
We set $s_{t,i}^{\text{Entropy}} = \mathcal{H}(b_{t-1,i})$, so high scores correspond to high uncertainty.
This variant approximates the information-gain objective by repeatedly testing ambiguous documents.

\noindent \textbf{\textsc{GP-Belief} (signal anchoring).}
We set $s_{t,i}^{\text{Belief}} = b_{t-1,i}$, so high scores correspond to documents believed most likely to be relevant.
This variant approximates the active search objective by repeatedly anchoring likely-relevant documents in high-diagnosticity positions.

We formalize both objectives in \S\ref{sec:Theory}, showing that uncertainty reduction is aligned with maximizing expected entropy decrease under our detector model, while signal anchoring maximizes a belief-weighted sum of position-wise detection rates.
The algorithm returns the final posterior beliefs $\mathbf{b}_T$, which can be mapped to discrete predictions via standard decision rules, as detailed in the experimental setup (\S\ref{subsec:Setup}).

\subsection{Computational Complexity}
\label{subsec:complexity_main}

The per-round complexity of \GP{} is $O(N \log N)$ for sorting documents by score, compared to $O(N)$ for PSC's random shuffle.
However, this is negligible CPU overhead, as the dominant cost in practice is LLM inference.
With respect to the number of costly LLM calls, \GP{} provides substantial savings, identifying targets in $O(\log N)$ rounds (proven in \S\ref{subsec:convergence}), compared to $O(T)$ rounds for PSC, where $T \gg \log N$ in practice.

\section{Theoretical Properties}
\label{sec:Theory}

We analyze \GP{} as a sequential Bayesian inference process.
Here we characterize how the rounds required to identify a relevant document scale with collection size and assigned position quality.
The central quantity governing convergence is the belief update \emph{drift}, determined by the diagnostic strength of the assigned positions.
All proofs are deferred to Appendix~\ref{app:proofs}.

\subsection{Belief Dynamics and Sample Complexity}
\label{subsec:convergence}

Recall that the log-odds evolve as a random walk with increments determined by the assigned position (\S\ref{subsec:beliefs}).
For a relevant document assigned to position $j$, the expected increment of this random walk is
\begin{equation}
    \mu_j
    =
    \mathbb{E}[\ell_j(O)\mid Z=1]
    =
    D_{\mathrm{KL}}(P_{Z=1,j}\,\|\,P_{Z=0,j}),
\end{equation}
the KL divergence between the observation distributions induced by position $j$.
Positions with larger $\mu_j$ therefore induce faster belief concentration.
The total rounds required to identify the relevant document thus depends on the cumulative drift achieved by the policy's position assignments.

\begin{theorem}[Sample Complexity]
\label{thm:sample_complexity}
Assume that log-likelihood ratios are bounded and that the policy assigns the relevant document positions with average drift at least $\underline{\mu} > 0$, i.e.,
$
    \sum_{t=1}^T \mu_{\sigma_t(i)} \ge \underline{\mu} \cdot T.
$
To identify a unique relevant document among $N$ candidates with high probability, the number of rounds required satisfies
\[
    T
    =
    O\!\left(
        \frac{\log N}{\underline{\mu}}
        +
        \frac{\log N}{\underline{\mu}^2}
    \right)
    =
    O\!\left(
        \frac{\log N}{\underline{\mu}^2}
    \right).
\]
\end{theorem}

\noindent\textbf{Interpretation.}
The $\log N$ factor reflects the intrinsic difficulty of distinguishing one relevant document from $N-1$ distractors.
The dominant term scales quadratically with $1/\underline{\mu}$, indicating that convergence speed is governed primarily by the diagnostic quality of the positions assigned to the true positive.
Policies that concentrate high-drift positions on promising candidates therefore achieve substantially faster identification.

\subsection{Information Rate Advantage of Strategic Placement}
\label{subsec:rate}
Thm~\ref{thm:sample_complexity} established that the convergence time is inversely proportional to the assigned drift $\mu$.
Consequently, the efficiency of the retrieval process depends entirely on the policy's ability to maximize this quantity.
We now quantify exactly how much \textsc{GP-Belief} improves the information rate compared to random assignment.

Let $\mu_{(1)} \ge \dots \ge \mu_{(N)}$ denote the position drifts sorted in descending order of quality.
Let $\mathrm{rank}_t(i) \in \{1, \dots, N\}$ be the rank of document $i$ when sorting current beliefs $b_{t-1,i}$ in descending order.
\begin{definition}[Information Rate]
    The \emph{information rate} for a relevant document $i$ under policy $\pi$ at round $t$ is the expected drift assigned to it, conditioned on the history:
    \[
        \mathcal{R}_\pi(t) = \mathbb{E} \left[ \mu_{\sigma_t(i)} \mid Z_i = 1, \mathcal{F}_{t-1} \right].
    \]
\end{definition}
For the baseline \textbf{Permutation Self-Consistency (PSC)}, which assigns documents uniformly at random, the rate is fixed at the average drift: $\mathcal{R}_{\mathrm{PSC}}(t) = \bar{\mu} = \frac{1}{N} \sum_{j=1}^N \mu_j$.
\begin{proposition}[Exact Rate for \textsc{GP-Belief}]
\label{prop:rate_exact}
    Under \textsc{GP-Belief}, the information rate is:
    \[
        \mathcal{R}_{\textsc{GP-Belief}}(t)
        \!=\!
        \sum_{r=1}^N \mu_{(r)}\!\cdot\!\Pr \left( \mathrm{rank}_t(i)\!=\!r\!\mid\!Z_i\!=\!1, \mathcal{F}_{t-1} \right).
    \]
\end{proposition}
To interpret this, we look at the probability of the relevant document being ``captured'' in the high-quality positions.
\begin{definition}[Top-$m$ Capture Probability]
    For any cut-off $m \in \{1, \dots, N\}$, let
    \[
        p_t(m) = \Pr \left( \mathrm{rank}_t(i) \le m \mid Z_i=1, \mathcal{F}_{t-1} \right),
    \]
    be the probability that the relevant document is ranked among the top $m$ candidates.
\end{definition}
\begin{proposition}[Rate Advantage]
\label{prop:rate_advantage}
    For any $m \in \{1,\dots,N\}$, the information rate of \textsc{GP-Belief} satisfies the lower bound:
    \[
        \mathcal{R}_{\textsc{GP-Belief}}(t)
        \ge
        p_t(m)\, \mu_{(m)} \;+\; (1-p_t(m))\, \mu_{(N)}.
    \]
\end{proposition}
\noindent \textbf{Interpretation.}
This bound isolates the two drivers of \GP{}'s performance:
(1) \textbf{Ranking Quality ($p_t(m)$):} As the algorithm progresses, the probability of the relevant document being in the top $m$ increases.
Under random guessing (PSC), this probability is simply $m/N$.
Whenever $p_t(m) > m/N$, \GP{} is guaranteed to outperform the baseline.
(2) \textbf{Position Heterogeneity ($\mu_{(m)} - \mu_{(N)}$):} The advantage is amplified when the position profile is ``spiky.''
If the top positions ($\mu_{(m)}$) are better than the worst ones ($\mu_{(N)}$), identifying the correct document accelerates in subsequent rounds.

This creates a positive feedback loop: a good ranking yields a high drift $\mu$, which accumulates evidence faster, which further improves the ranking $p_{t+1}(m)$.

\subsection{Why Anchoring Dominates Uncertainty Reduction}
\label{subsec:anchoring_vs_entropy}

Prop.~\ref{prop:rate_exact}--\ref{prop:rate_advantage} establish that assigning documents to high-drift positions accelerates convergence.
The central question remains: \emph{which} documents should receive these positions?

\textsc{GP-Entropy} prioritizes high-uncertainty documents (maximizing information gain), while \textsc{GP-Belief} prioritizes high-belief documents (maximizing expected discoveries).
In sparse search settings ($k \ll N$), these objectives diverge sharply.
We formalize this divergence by analyzing the drift allocated to the true positive as it emerges from the haystack.

\begin{proposition}[\textsc{GP-Entropy} Demotes Confident Positives]
\label{prop:entropy_demotion}
    Consider the $k=1$ setting.
    As the belief in the true positive approaches certainty ($b_{t-1,i} \to 1$), its entropy $H(b_{t-1,i}) \to 0$.
    Consequently, if there exist at least $m$ distractor documents with higher entropy (e.g., $b \approx 0.5$), \textsc{GP-Entropy} assigns the true positive a rank greater than $m$.
    In the limit, the diagnostic attention collapses to the minimum available drift:
    \[
        \lim_{b_{t-1,i} \to 1} \mathcal{R}_{\textsc{GP-Entropy}}(t) = \mu_{(N)}.
    \]
\end{proposition}

\begin{figure*}[t]
    \centering
    \scalebox{1}[1]{
      \includegraphics[width=\linewidth,trim=0.0cm 0.3cm 0cm 0.2cm]{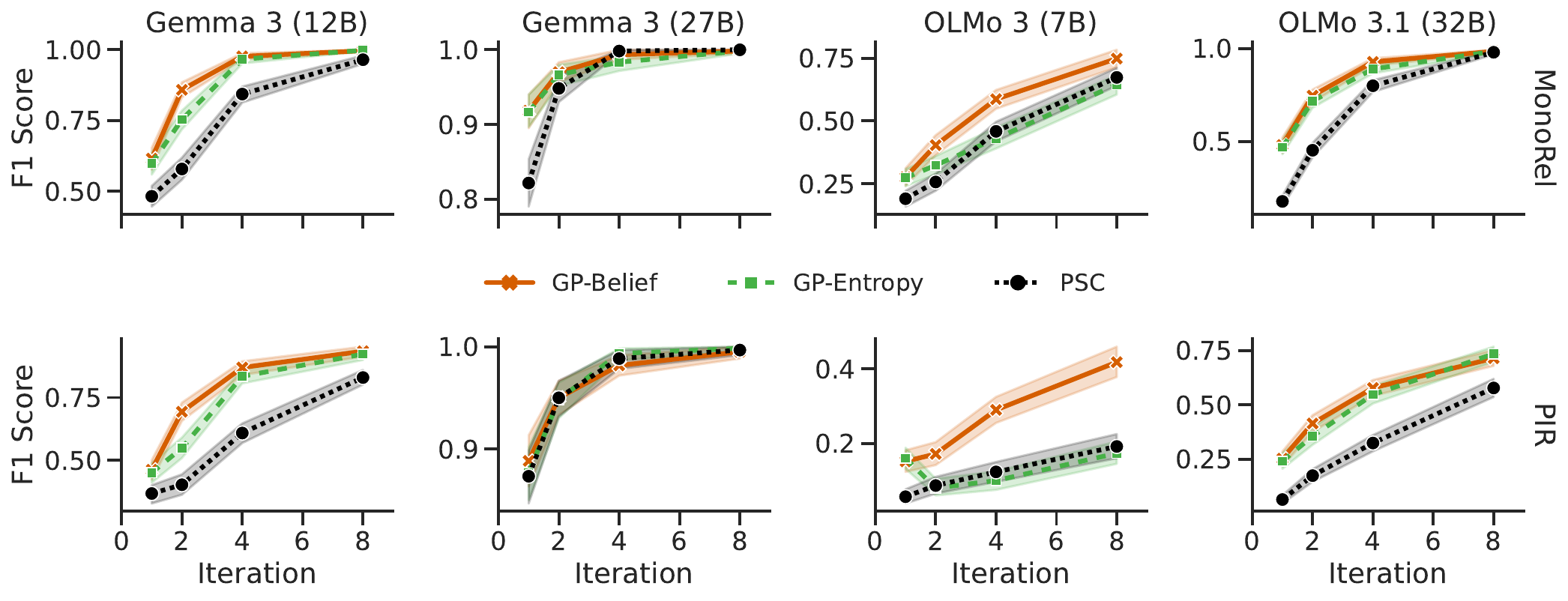}
    }
    \caption{
        \textbf{Main results across models and tasks.}
        $F_1$ score over 8 iterations for $N{=}100$ documents.
        \textsc{GP-Belief} (orange) consistently outperforms \textsc{GP-Entropy} (green, dashed) and PSC (black, dotted), with advantages most pronounced on models exhibiting strong position bias.
        Gemma-3-27B shows no separation due to its lack of exploitable bias.
        Shaded regions indicate 95\% confidence intervals.
    }
    \label{fig:main_results}
\end{figure*}

\begin{proposition}[\textsc{GP-Belief} Diagnostic Lock-in]
\label{prop:belief_lockin}
    Consider the $k=1$ setting.
    Under \textsc{GP-Belief}, if the true positive $i$ achieves the highest belief rank at round $\tau$, it receives maximal diagnostic attention $\mu_{(1)}$.
    Furthermore, there exists a strictly positive probability that $i$ retains the lead in all subsequent rounds $t > \tau$, maintaining $\mathcal{R}_{\textsc{GP-Belief}}(t) = \mu_{(1)}$.
\end{proposition}
\noindent \textbf{Interpretation: Vicious vs. Virtuous Cycles.}
\textbf{The Vicious Cycle of Entropy:} \textsc{GP-Entropy} penalizes success.
As the model begins to identify the relevant document (belief rises), its entropy falls.
The policy effectively says, ``We are sure about this one, so let's stop checking it.''
It demotes the candidate to the noisiest positions ($\mu_{(N)}$), starving it of the signal required to cross the final confidence threshold ($1-\delta$). 
This explains the stagnation observed in Figure~\ref{fig:main_results}.
\textbf{The Virtuous Cycle of Belief:} \textsc{GP-Belief} rewards success.
As the relevant document rises in rank, it is promoted to positions with higher drift ($\mu_{(1)}$).
This strong signal reinforces the belief, further solidifying its top rank.
This ``Signal Anchoring'' creates a lock-in effect that races the winner to the threshold with maximal velocity.

\section{Empirical Validation}
\label{sec:Experiments}

\subsection{Setup}
\label{subsec:Setup}

\textbf{Benchmarks.}
We evaluate on two FLenQA benchmarks~\citep{levy2024sametask}: \textbf{MonoRel} (monotonic transitive reasoning) and \textbf{PIR} (People in Rooms; compositional reasoning).
We selected these tasks because they offer unambiguous, deterministic ground-truth labels.
In contrast to open-domain QA, where implicit ambiguity can confound evaluation~\citep{min2020ambigqa} and necessitate complex semantic matching~\citep{kamalloo2023evaluating}, FLenQA allows for precise measurement of retrieval recall without the noise of partial relevance or answer granularity mismatches.
While these benchmarks are natively single-fact QA, we convert each instance into a controlled MDQA task by augmenting the ground-truth fact with $N-1$ sampled distractor, yielding exactly one relevant document.

\textbf{Models.}
We test two model families at multiple scales: \textbf{Gemma-3}~\citep{gemma3} (12B, 27B) and \textbf{OLMo-3}~\citep{olmo3} (OLMo-3-7B, OLMo-3.1-32B).
These models exhibit varying degrees of position bias as measured by their calibrated positional profiles (\S\ref{app:calibration}), allowing us to isolate when position-aware reordering helps.

\newpage

\textbf{Default configuration.}
Unless stated otherwise, we use $N = 100$ documents with $k = 1$ relevant document (Top-$1$ selection) and run $T = 8$ rounds.
We follow model-recommended sampling parameters: temperature $1.0$ with top-$p$ $0.95$ for Gemma, and temperature $0.6$ with top-$p$ $0.95$ for OLMo.
All methods enforce structured JSON outputs via constrained decoding to ensure parsable citations.
Each round returns a set of cited document IDs, which we map to per-document indicators $O_{t,i}$ (Eq.~\eqref{eq:observation_model}) for belief updates.

\textbf{Baseline.}
We compare against \textbf{Permutation Self-Consistency (PSC)}~\citep{tang2024found}, which queries the model on independent random permutations and aggregates per-document relevance predictions across rounds via per-document majority vote (ties broken deterministically).
PSC is a black-box baseline that mitigates position bias by averaging over placements.
All methods use the same number of LLM calls, enabling direct comparison of query efficiency.

\textbf{Evaluation.}
\GP{} maintains continuous beliefs $\mathbf{b}_t \in [0,1]^N$, converted to binary predictions for evaluation.
We assume the number of relevant documents $k$ is known and given, and select the $k$ highest-belief documents (Top-$k$).
We report mean $F_1$ with 95\% confidence interval.

\textbf{Calibration of diagnosticity.}
For each model, we run a lightweight calibration to extract its diagnosticity profile.
At position $j$, we construct calibration trials by placing a known-relevant ``gold'' document at position $j$, filling the remaining $N-1$ positions with distractors, and query the model.
Citing the gold at $j$ contributes to $\mathrm{TPR}_j$, while citing position $j$ when gold is elsewhere contributes to $\mathrm{FPR}_j$.
Estimating diagnosticity at \textit{all} $N$ positions can be expensive for long contexts, even though this is a one-time cost.
We therefore use sparse fixed-grid sampling with linear interpolation, assuming smooth diagnosticity trends.
Specifically, we evaluate a fixed grid of $K=11$ positions with 50 trials each and linearly interpolate to obtain the full $N$-position profile (validated in \S\ref{subsec:Calibration}; full details in Appendix~\ref{app:calibration}).

Fig.~\ref{fig:diagnosticity_profile} shows the calibrated diagnosticity profile for Gemma-3-12B on both MonoRel and PIR with $N=100$ documents.
Both exhibit the characteristic ``lost in the middle'' pattern: positions near the beginning (primacy) and end (recency) of context show high diagnosticity, while middle positions are far less informative.
This heterogeneity enables strategic placement---\textsc{GP-Belief} exploits the high-diagnosticity positions to accelerate belief separation.

\begin{figure}[h]
    \centering
    \includegraphics[width=0.85\columnwidth,trim=0.5cm 0.3cm 0cm 0.2cm]{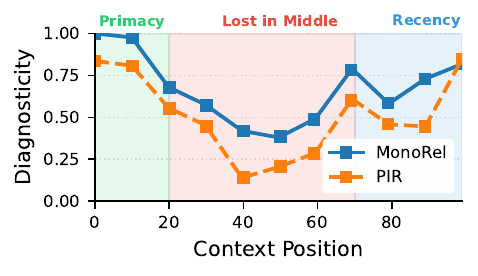}
    \caption{\textbf{Diagnosticity profile for Gemma-3-12B on MonoRel and PIR ($N=100$).} 
    Both tasks show primacy and recency effects, i.e., diminished diagnosticity in  the middle. 
    }
    \label{fig:diagnosticity_profile}
\end{figure}

\subsection{Main Results}
\label{subsec:MainResults}

Fig.~\ref{fig:main_results} compares \textsc{GP-Belief}, \textsc{GP-Entropy}, and PSC across four models.
We analyze whether (i) our stateful Bayesian inference beats stateless ensembling, and (ii) signal anchoring outperforms uncertainty reduction.

\textbf{Signal anchoring (\textsc{GP-Belief}) dominates uncertainty reduction (\textsc{GP-Entropy}).}
Whenever position bias is exploitable, \textsc{GP-Belief} consistently outperforms \textsc{GP-Entropy}.
On Gemma-3-12B (PIR), \textsc{GP-Belief} achieves $F_1 \approx 0.88$ at iteration 4 versus $0.72$ for \textsc{GP-Entropy}, a gap that persists across model scales. 
This confirms that in needle-in-haystack retrieval ($k \ll N$), the optimal strategy is not to probe ambiguity, but to \emph{protect likely signals}.
\textsc{GP-Entropy} fails because it prioritizes documents near $b \approx 0.5$---predominantly irrelevant distractors not yet ruled out.
By contrast, \textsc{GP-Belief} \emph{anchors likely signals} in high-diagnosticity positions to accelerate separation.

\textbf{Bayesian statefulness beats stateless ensembling.}
Even with the suboptimal entropy objective, the Bayesian framework generally outperforms the stateless PSC baseline.
On OLMo-3-7B (PIR), \textsc{GP-Entropy} attains $F_1 \approx 0.25$ at iteration 8 compared to PSC's $0.17$.
The gains come from \emph{adaptivity}: PSC treats iterations independently, often re-testing resolved negatives in high-value positions, whereas \GP{} accumulates evidence, deprioritizing low-belief documents to make room for promising ones.
When combined with \textsc{GP-Belief}, this statefulness enables the theoretical information rate gains of Proposition~\ref{prop:rate_advantage}.

\textbf{Query efficiency.}
\textsc{GP-Belief} matches PSC using \textbf{30\text{--}65\% fewer queries}.
On Gemma-3-12B (MonoRel), \textsc{GP-Belief} reaches $F_1 \approx 0.97$ in just 4 iterations, matching PSC's performance at iteration 8.
On PIR, the gains are even larger: \textsc{GP-Belief}'s iteration-4 performance ($0.88$) strictly exceeds PSC's final iteration-8 result ($0.82$), making it a viable strategy for latency-constrained applications.

\textbf{No bias, no gains.}
Gemma-3-27B (Fig.~\ref{fig:main_results}, column 2) serves as a control.
Because this model exhibits a fairly flat diagnosticity profile (Fig.~\ref{fig:calibration_facet_N100}), strategic reordering offers no leverage over random shuffling, and all methods converge indistinguishably.
This negative result validates that \GP{} is not a ``magic wand,'' but a targeted intervention that turns position bias into a resource.

\subsection{Validating Sparse Diagnosticity Calibration}
\label{subsec:Calibration}

As discussed in \S\ref{subsec:Setup}, we estimate diagnosticity via sparse position sampling.
We validate this by comparing fixed-grid interpolation against uniform calibration over \textit{all} positions.
Tab.~\ref{tab:sampling_validation} reports the mean absolute residual $\Delta$ between fixed-grid interpolated values and uniformly-sampled ground truth ($N=100$); \textbf{across models, $\Delta < 0.071$.}
Thus, a grid of $K=11$ positions recovers diagnosticity with minimal bias, \textit{supporting lightweight calibration for full-scale evaluation.}

\begin{table}[ht]
    \centering
    \footnotesize
    \caption{Mean absolute residual $\Delta$ (mean$_{\pm \text{SE}}$) between uniformly-sampled ($N=100$) and fixed-grid interpolated diagnosticity ($K = 11$ positions).
    All values $< 0.071$ confirm that sparse grid sampling  accurately recovers the full profile.}
    \begin{tabular}{c l c }
        \toprule
        \multirow{1}{*}{Dataset} & \multirow{1}{*}{Model} & Mean $\Delta$ Diagnosticity \\
        \midrule
        \multirow{4}{*}{\rotatebox[origin=c]{90}{\textbf{MonoRel}}}
         & Olmo-3-7B-Think & $0.054_{\pm 0.002}$ \\
         & Olmo-3.1-32B-Think & $0.051_{\pm 0.002}$ \\         
         & Gemma-3-12B-IT & $0.027_{\pm 0.002}$ \\
         & Gemma-3-27B-IT & $0.008_{\pm 0.000}$ \\
        \midrule
        \multirow{4}{*}{\rotatebox[origin=c]{90}{\textbf{PIR}}}
         & Olmo-3-7B-Think & $0.041_{\pm 0.002}$ \\
         & Olmo-3.1-32B-Think & $0.071_{\pm 0.003}$ \\
         & Gemma-3-12B-IT & $0.028_{\pm 0.002}$ \\
         & Gemma-3-27B-IT & $0.013_{\pm 0.001}$ \\
        \bottomrule
    \end{tabular}
    \label{tab:sampling_validation}
\end{table}

\subsection{Cross-dataset Transferability}
\label{subsec:Transferability}

\begin{figure}[ht]
    \centering
    \includegraphics[width=0.95\linewidth,trim=0.0cm 0.3cm 0cm 0.2cm]{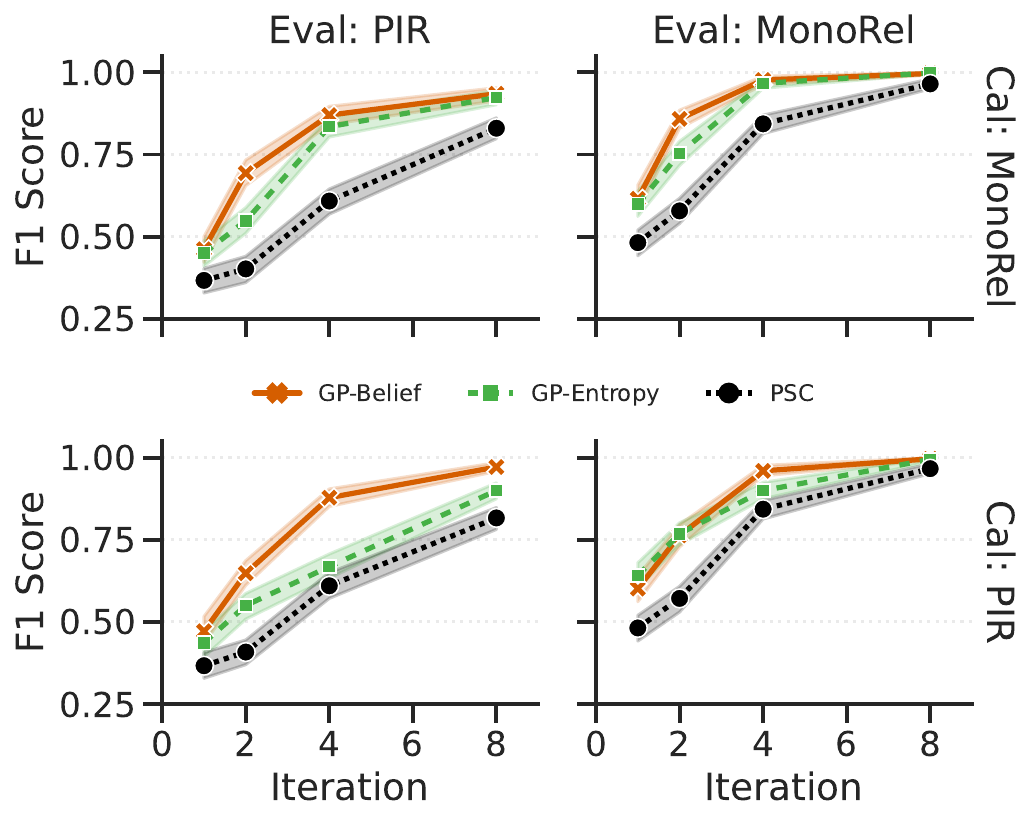}
    \caption{\textbf{Cross-task transferability for Gemma-3-12B ($N=100$).} 
Rows indicate calibration source; columns indicate evaluation task.
\textsc{GP-Belief} maintains its advantage even under cross-calibration (off-diagonal), with only modest degradation compared to matched calibration (diagonal).
This robustness suggests that coarse positional ordering, the relative ranking of high- vs. low-diagnosticity positions, drives most of the gain.}
    \label{fig:transfer}
\end{figure}

If position bias is a structural model property (as suggested by Fig.~\ref{fig:diagnosticity_profile}) rather than dataset-specific, diagnosticity profiles should transfer across datasets.
We test this via cross-evaluation: 
\textit{applying the diagnosticity profile calibrated on MonoRel to guide GP on PIR, and vice-versa.}
As shown in Fig.~\ref{fig:transfer}, \textbf{\textsc{GP-Belief} retains its advantage under cross-calibration}, with only modest degradation relative to matched calibration.
This suggests most gains come from  \emph{coarse positional ordering}, i.e., with the ranking of positions being more important than precise TPR/FPR values.

\begin{figure*}
    \centering
    \includegraphics[width=0.95\linewidth,trim=0.0cm 0.3cm 0cm 0.2cm]{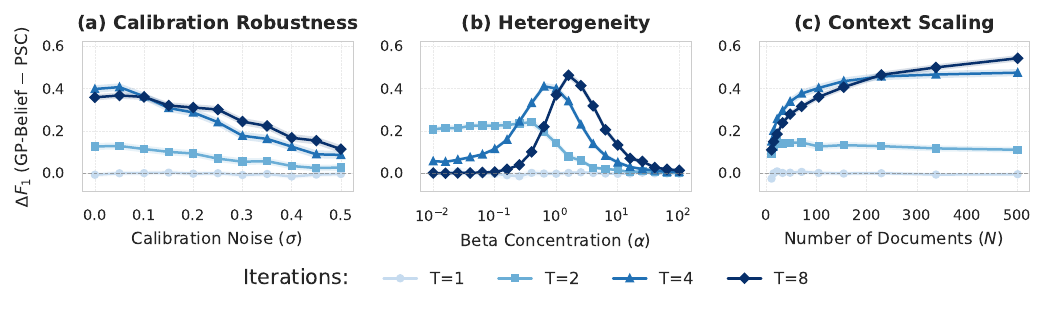}
    \caption{\textbf{Ablation studies characterizing \textsc{GP-Belief}'s operating conditions (simulation; Top-1 selection, N=100, k=1, 5,000 trials unless noted).}
(a) Calibration robustness: advantage degrades gracefully as Gaussian noise ($\sigma$) is injected into TPR/FPR estimates, confirming that relative position rankings matter more than precise calibration.
(b) Heterogeneity: advantage peaks at moderate Beta concentration ($\alpha \approx 1$) and vanishes as positions become homogeneous ($\alpha \to \infty$), since uniform detectors offer no strategic leverage.
(c) Context scaling: advantage grows with $N$, reaching $\Delta F_1 \approx 0.5$ at $N=500$, confirming that strategic placement becomes critical as random placement increasingly misses high-diagnosticity positions.}
    \label{fig:ablations}
\end{figure*}

\subsection{Controlled Ablations with Simulated Environment}
\label{subsec:Ablations}

Real LLM experiments confound multiple factors: calibration error, intrinsic model variability, and task-specific noise.
To isolate the conditions under which \textsc{GP-Belief} succeeds, we turn to simulation, where we can precisely control detector heterogeneity, calibration accuracy, and context scale independently.
We characterize \textsc{GP-Belief}'s operating conditions through simulation (Fig.~\ref{fig:ablations}; Top-1 selection, $N=100$, $k=1$, $5\,000$ trials unless noted).

\textbf{Robustness to calibration error (a).} We test \textsc{GP-Belief} under miscalibration by adding Gaussian noise ($\sigma \in [0, 0.5]$) to TPR/FPR estimates (Fig.~\ref{fig:ablations}a).
Performance degrades smoothly: even at $\sigma = 0.4$, \textsc{GP-Belief} retains a clear advantage over PSC ($\Delta F_1 \approx 0.15\text{--}0.25$).
This indicates the method relies on relative ranking of positional diagnosticities rather than their precise absolute values.

\textbf{Heterogeneity (b).} \textsc{GP-Belief}'s advantage depends on profile variation.
When detector quality is drawn from \textsc{Beta}($\alpha$, $\alpha$), advantage peaks at moderate concentration ($\alpha \approx 1$) and vanishes as positions become homogeneous ($\alpha \to \infty$), since uniform detectors are indistinguishable and therefore offer no opportunity for strategic leverage.

\textbf{Context scaling (c).} The advantage grows with $N$, reaching $\Delta F_1 \approx 0.5$ at $N = 500$.
As context expands, random placement rarely assigns relevant documents to diagnostic positions, amplifying the value of strategic placement.

\section{Related Work}
\label{sec:RelatedWork}

\noindent\textbf{Exploiting systematic biases.}
Systematic biases are typically viewed as obstacles, but prior work in human--computer interaction has leveraged cognitive biases to optimize interfaces~\citep{gajos2005preference, gajos2008improving} and guide user behavior~\citep{cialdini2007influence, mathur2019dark}.
In LLMs, predictable regularities enable similar efficiency gains, as in speculative decoding and early-exit methods~\citep{leviathan2023fast, schuster2022confident, chen2024eellm}.
In a similar spirit, we leverage position bias as a \emph{diagnostic signal} to improve in-context NIAH retrieval.

\noindent\textbf{Position bias in LLMs.}
Position bias in long-context LLMs is well-studied~\citep[inter alia]{wang2023primacy, zheng2023large, liu2024lost}, with mitigations broadly divided into white-box and black-box approaches.
\emph{White-box methods} suppress bias by modifying model internals, including alternative positional encodings~\citep{zhang2024found} and attention manipulations~\citep{peysakhovich2023attention, hsieh2024found, wang2024eliminating}.
\emph{Black-box methods} instead treat the model as fixed, such as
Permutation Self-Consistency (PSC)~\citep{tang2024found}, which averages predictions over random permutations.
PSC, however, is stateless and does not exploit information accumulated across queries.
While \citet{liu2025selfelicit} explores relevance via internal model states, we are the first to formalize inference-time reordering as \emph{sequential Bayesian active search}, enabling stateful inference using only API-level outputs.

\noindent\textbf{Active search and iterative retrieval.}
We frame context reordering as a \emph{Bayesian active search} problem~\citep{garnett2012bayesian, jiang2017efficient}, distinct from conventional active learning~\citep{settles2009active, settles2011theories}, which prioritizes uncertainty reduction.
Bandit-based methods~\citep{slivkins2019introduction} have been applied to optimize retrieval in RAG systems~\citep{duan2025chunks}, but these primarily operate at training time.
At inference time, \GP{} is most closely related to iterative reranking methods such as RankGPT~\citep{sun2023chatgpt}, but differs fundamentally: rather than prompting the model to explicitly sort candidates (itself subject to position bias), we infer relevance from the model's natural responses under structured layouts, requiring no instruction tuning.

\section{Conclusion}
\label{sec:Conclusion}

We introduce \GP{}, a Bayesian framework that exploits position bias as a diagnostic signal for long-context retrieval.
By treating positions as noisy detectors and actively anchoring high-belief documents in high-diagnosticity positions, signal anchoring consistently outperforms uncertainty-based baselines, achieving large query reductions with matched asymptotic accuracy.
More broadly, our results suggest that predictable inference-time biases can be leveraged as information rather than treated as noise.






\section*{Impact Statement}

This paper presents a method for improving the efficiency of long-context retrieval in large language models.
By reducing the number of LLM queries required to identify relevant documents, \GP{} may lower computational costs and latency in retrieval-augmented generation systems, with associated reductions in energy consumption.

Our method exploits position bias, a known limitation of current LLMs.
While we frame this bias as a resource rather than a flaw, we do not believe this work discourages efforts to reduce position bias at the model level.
We see no direct ethical concerns specific to this work beyond those inherent to improving information retrieval systems generally.

\bibliography{icml2026_custom, ref}
\bibliographystyle{icml2026}

\newpage
\appendix

\section{Proofs}
\label{app:proofs}

In this section, we provide proofs for the theoretical properties of \GP{}.
We begin by defining the filtration and the martingale structure of the belief updates.

\textbf{Setup and notation.}
Let $(\Omega, \mathcal{F}, P)$ be a probability space carrying the sequence of document relevance variables $Z$ and observation sequence $O$.
Let $\mathcal{F}_t = \sigma(O_1, \dots, O_t, \sigma_1, \dots, \sigma_{t+1})$ be the filtration representing the information available after round $t$ (including the policy's choice for round $t+1$).
The log-odds for a fixed document $i$ update as:
\[
    \lambda_{t,i} = \lambda_{t-1,i} + \ell_{\sigma_t(i)}(O_{t,i}),
\]
where $\ell_j(o)$ is the log-likelihood ratio at position $j$.

\subsection{Proof of Consistency}

\begin{proof}
Consider a single document $i$.
Assume without loss of generality that $i$ is relevant ($Z_i=1$); the case for $Z_i=0$ follows by symmetry with negative drift.
Let $X_t = \ell_{\sigma_t(i)}(O_{t,i})$ denote the update increment at round $t$.
We can decompose the accumulated log-odds $\lambda_{t,i} = \sum_{s=1}^t X_s$ into a drift term and a martingale difference sequence.

Define the expected drift at step $s$ conditioned on the history:
\[
    \mu_s = \mathbb{E}[X_s \mid \mathcal{F}_{s-1}, Z_i=1].
\]
By definition of the log-likelihood ratio, this expectation is the KL divergence:
\[
    \mu_s = D_{\mathrm{KL}}(P_{Z=1, \sigma_s(i)} \| P_{Z=0, \sigma_s(i)}).
\]
Let $M_t$ be the centered process:
\[
    M_t = \sum_{s=1}^t (X_s - \mu_s).
\]
$M_t$ is a martingale with respect to $\mathcal{F}_t$ because $\mathbb{E}[M_t - M_{t-1} \mid \mathcal{F}_{t-1}] = \mathbb{E}[X_t - \mu_t \mid \mathcal{F}_{t-1}] = 0$.

As in Thm.~\ref{thm:sample_complexity}, we assume the drift is bounded, and so $|X_s| \le L$.
Consequently, the martingale increments are also bounded: $|X_s - \mu_s| \le 2L$.
By the Strong Law of Large Numbers for Martingales, we have:
\[
    \lim_{t \to \infty} \frac{M_t}{t} = 0 \quad \text{almost surely}.
\]

Now consider the log-odds normalized by time:
\[
    \frac{\lambda_{t,i}}{t} = \frac{1}{t} \sum_{s=1}^t \mu_s + \frac{M_t}{t}.
\]
As $t \to \infty$, the second term vanishes.
For the first term, the boundedness guarantees that the average drift is bounded away from zero:
\[
    \liminf_{t \to \infty} \frac{1}{t} \sum_{s=1}^t \mu_s \ge \underline{\mu} > 0.
\]
Therefore, $\liminf_{t \to \infty} \frac{\lambda_{t,i}}{t} \ge \underline{\mu} > 0$.
This implies $\lim_{t \to \infty} \lambda_{t,i} = +\infty$ almost surely.
Since the belief $b_{t,i} = \sigma(\lambda_{t,i})$ is the sigmoid of the log-odds, $b_{t,i} \xrightarrow{a.s.} 1$, recovering the ground truth $Z_i=1$.
\end{proof}

\subsection{Proof of Thm.~\ref{thm:sample_complexity} (Sample Complexity)}

\begin{proof}
We analyze the probability that the log-odds $\lambda_{T,i}$ fail to reach the threshold $\Lambda$ by time $T$; assume $Z_i=1$.
We are given that for all $t \le T$, the assigned positions satisfy $\mu_{\sigma_t(i)} \ge \underline{\mu}$.
Decomposing the log-odds as before:
\[
    \lambda_{T,i} = \sum_{t=1}^T \mu_t + M_T \ge T\underline{\mu} + M_T.
\]

A failure occurs if $\lambda_{T,i} < \Lambda$.
This implies:
\[
    T\underline{\mu} + M_T < \Lambda \implies M_T < \Lambda - T\underline{\mu}.
\]

We apply the Azuma-Hoeffding inequality to the martingale $M_T$.
The increments $Y_t = X_t - \mu_t$ are bounded within an interval of size at most $2L$ (since $X_t \in [-L, L]$).
The inequality states that for any $K > 0$:
\[
    \Pr(M_T \le -K) \le \exp\left( -\frac{K^2}{2TL^2} \right).
\]

Let us choose $T$ large enough such that the expected drift $T\underline{\mu}$ exceeds the threshold $\Lambda$ by a safety margin $K$.
Specifically, we set the safety margin to be half the total drift:
\[
    \text{Let } T\underline{\mu} = 2\Lambda \implies \Lambda - T\underline{\mu} = -\Lambda.
\]

However, we also need to satisfy the confidence requirement $\delta'$.
We set $T$ according to the bound in the proposition:
\[
    T = \frac{2\Lambda}{\underline{\mu}} + \frac{8L^2}{\underline{\mu}^2} \log\left(\frac{1}{\delta'}\right).
\]

Substituting this $T$ into the failure condition is algebraically complex, so we verify the bound using the safety margin approach directly.
Let us define the target margin $K = T\underline{\mu} - \Lambda$.
We require $\Pr(M_T < -K) \le \delta'$.
Using Azuma-Hoeffding, it suffices to have:
\begin{align*}
    \exp\left( -\frac{K^2}{2TL^2} \right) \le \delta'
     &\implies \frac{K^2}{2TL^2} \ge \log(1/\delta') \\
     &\implies K \ge L\sqrt{2T \log(1/\delta')}.
\end{align*}

Substitute $K = T\underline{\mu} - \Lambda$:
\[
    T\underline{\mu} - \Lambda \ge L\sqrt{2T \log(1/\delta')}.
\]

We now check if the proposed $T$ satisfies this inequality.
The proposed $T$ has two terms.
Let $T_1 = \frac{2\Lambda}{\underline{\mu}}$ and $T_2 = \frac{8L^2}{\underline{\mu}^2} \log(1/\delta')$.
So $T = T_1 + T_2$.
The LHS of the inequality becomes:
\[
    (T_1 + T_2)\underline{\mu} - \Lambda = (2\Lambda + T_2\underline{\mu}) - \Lambda = \Lambda + T_2\underline{\mu} > T_2\underline{\mu}.
\]

The RHS requires us to bound $L\sqrt{2(T_1+T_2) \log(1/\delta')}$.
Notice that for $\delta'$ small enough (or $\Lambda$ small relative to the noise), the stochastic term dominates.
To strictly prove the bound holds, we can simply ensure $T_2\underline{\mu}$ alone is sufficient to cover the noise from the \emph{entire} duration $T$.
Observe that $T_2\underline{\mu} = \frac{8L^2}{\underline{\mu}} \log(1/\delta')$.
Squaring the LHS requirement (conservatively using just $T_2\underline{\mu}$):
\[
    (T_2\underline{\mu})^2 = \frac{64 L^4}{\underline{\mu}^2} (\log(1/\delta'))^2.
\]

Squaring the RHS (substituting $T \approx T_2$ for the worst case noise regime):
\[
    2 T_2 L^2 \log(1/\delta') = \frac{16 L^4}{\underline{\mu}^2} (\log(1/\delta'))^2.
\]

Since $64 > 16$, the drift from $T_2$ alone is more than sufficient to cover the variance from $T_2$.
The addition of $T_1$ adds more drift ($\Lambda$) than variance (since variance grows as $\sqrt{T}$ while drift grows as $T$), so the inequality holds comfortably.
Thus, with probability at least $1-\delta'$, the threshold is crossed by time $T$.

We note that the derived bound for $T$ is an upper bound and is not tight.
The constants $2$ and $8$ are conservative estimates chosen to simplify the application of the Azuma-Hoeffding inequality and ensure the safety margin $K$ is robust against worst-case noise.
While a tighter constant might be achievable through more complex analysis (e.g., using exact characteristic functions), the current bound is sufficient to establish the fundamental logarithmic scaling, $T \propto \log N$, required for efficient search.
\end{proof}

\subsection{Proofs of Information Rate Properties (Section~\ref{subsec:rate})}

\textbf{Setup.}
Recall that $\mu_{(1)} \ge \mu_{(2)} \ge \dots \ge \mu_{(N)}$ denotes the fixed position drifts sorted in descending order.
Let $\text{rank}_t(i)$ denote the rank of document $i$ according to the beliefs $b_{t-1}$ at the start of round $t$.
The \textsc{GP-Belief} policy assigns the document with rank $r$ to the position with drift $\mu_{(r)}$.
Thus, if $\text{rank}_t(i) = r$, the drift assigned to document $i$ is $\mu_{\sigma_t(i)} = \mu_{(r)}$.

\subsubsection{Proof of Proposition~\ref{prop:rate_exact} (Exact Rate)}

\begin{proof}
The information rate is defined as the expected drift assigned to a relevant document $i$ ($Z_i=1$) conditioned on the history $\mathcal{F}_{t-1}$:
\[
    \mathcal{R}_{\textsc{GP-Belief}}(t) = \mathbb{E} \left[ \mu_{\sigma_t(i)} \mid Z_i = 1, \mathcal{F}_{t-1} \right].
\]
Since the policy is deterministic given the beliefs (which are fixed in $\mathcal{F}_{t-1}$), the only randomness comes from the uncertainty of the rank itself relative to the ground truth.
Conditioning on the specific rank $r$ that the document achieves:
\[
    \mathbb{E} \left[ \mu_{\sigma_t(i)} \right] = \sum_{r=1}^N \mathbb{E} \left[ \mu_{\sigma_t(i)} \mid \text{rank}_t(i) = r \right] \cdot \Pr(\text{rank}_t(i) = r).
\]
Under \textsc{GP-Belief}, the assignment is strictly coupled: the event $\{\text{rank}_t(i) = r\}$ implies that the document is assigned to the position with drift $\mu_{(r)}$. Therefore, $\mathbb{E} \left[ \mu_{\sigma_t(i)} \mid \text{rank}_t(i) = r \right] = \mu_{(r)}$.
Substituting this back into the summation yields the exact rate:
\[
    \mathcal{R}_{\textsc{GP-Belief}}(t) = \sum_{r=1}^N \mu_{(r)} \cdot \Pr(\text{rank}_t(i) = r \mid Z_i=1, \mathcal{F}_{t-1}).
\]
\end{proof}

\subsubsection{Proof of Proposition~\ref{prop:rate_advantage} (Rate Advantage)}

\begin{proof}
We seek a lower bound on $\mathcal{R}_{\textsc{GP-Belief}}(t)$ based on the Top-$m$ capture probability $p_t(m)$.
We split the summation from Proposition~\ref{prop:rate_exact} into two parts: the top $m$ ranks (where the document is "captured") and the remaining $N-m$ ranks.
\begin{align*}
    \mathcal{R}_{\textsc{GP-Belief}}(t)
     &= \underbrace{\sum_{r=1}^m \mu_{(r)} \Pr(\text{rank}_t(i) = r)}_{\text{Term 1}} \\
     &+ \underbrace{\sum_{r=m+1}^N \mu_{(r)} \Pr(\text{rank}_t(i) = r)}_{\text{Term 2}}.
\end{align*}

\textbf{Bounding Term 1:}
The drifts are sorted descending, so for any rank $r \le m$, we have $\mu_{(r)} \ge \mu_{(m)}$.
\begin{align*}
    \text{Term 1}
     &\ge \sum_{r=1}^m \mu_{(m)} \Pr(\text{rank}_t(i) = r) \\
     &= \mu_{(m)} \sum_{r=1}^m \Pr(\text{rank}_t(i) = r).
\end{align*}
By definition, the sum of probabilities for the top $m$ ranks is the capture probability $p_t(m)$.
Thus, $\text{Term 1} \ge \mu_{(m)} p_t(m)$.

\textbf{Bounding Term 2:}
For any rank $r > m$, the drift is at least the minimum drift $\mu_{(N)}$.
\begin{align*}
    \text{Term 2}
     &\ge \sum_{r=m+1}^N \mu_{(N)} \Pr(\text{rank}_t(i) = r) \\
     &= \mu_{(N)} \sum_{r=m+1}^N \Pr(\text{rank}_t(i) = r).
\end{align*}
The sum of probabilities for the remaining ranks is simply the complement of the capture probability, $1 - p_t(m)$.
Thus, $\text{Term 2} \ge \mu_{(N)} (1 - p_t(m))$.

\textbf{Combining:}
Adding the lower bounds for Term 1 and Term 2 yields the result:
\[
    \mathcal{R}_{\textsc{GP-Belief}}(t) \ge \mu_{(m)} p_t(m) + \mu_{(N)} (1 - p_t(m)).
\]
\end{proof}

\subsection{Proofs of Signal Anchoring Properties (Section~\ref{subsec:anchoring_vs_entropy})}

\subsubsection{Proof of Proposition~\ref{prop:entropy_demotion} (Entropy Demotion)}

\begin{proof}
    The binary entropy function $H(p) = -p \log p - (1-p) \log (1-p)$ is concave and maximized at $p=0.5$.
    Consider the case where the true positive $i$ has high belief $b_{t-1,i} > 0.5$, and there are $N-1$ distractors.
    In sparse search ($k \ll N$), the vast majority of distractors are true negatives.
    However, early in the process, many distractors will have beliefs near the prior $0.5$ (high entropy) simply due to noise or lack of observation.

    Let $\mathcal{U} = \{j \neq i : H(b_{t-1,j}) > H(b_{t-1,i})\}$ be the set of documents with higher entropy than the true positive.
    \textsc{GP-Entropy} assigns positions by descending order of entropy.
    Thus, the rank of document $i$ is $|\mathcal{U}| + 1$.
    As $b_{t-1,i} \to 1$, $H(b_{t-1,i}) \to 0$.
    Since distractor beliefs fluctuate around $0.5$ (or decay to 0 slower than the winner rises to 1), the size of the set $|\mathcal{U}|$ approaches $N-1$ (all other documents have more uncertainty than the almost-certain winner).
    
    Consequently, the rank of $i$ approaches $N$.
    Under the assignment policy, rank $N$ corresponds to the position with drift $\mu_{(N)}$.
    Thus, $\mathcal{A}_{\textsc{GP-Entropy}}(t) \to \mu_{(N)}$.
\end{proof}

\subsubsection{Proof of Proposition~\ref{prop:belief_lockin} (Belief Lock-in)}

\begin{proof}
    Let $i$ be the relevant document ($Z_i=1$) and let $\mathcal{K} = \{k : k \neq i\}$ be the set of irrelevant distractors ($Z_k=0$).
    At round $\tau$, we are given that document $i$ achieves the highest belief rank.
    In terms of log-odds, this implies that $\lambda_{\tau,i} > \max_{k \in \mathcal{K}} \lambda_{\tau,k}$.

    Define the ``margin'' of the relevant document as the gap between its log-odds and that of its strongest competitor:
    \begin{equation}
        \Delta_t = \lambda_{t,i} - \max_{k \in \mathcal{K}} \lambda_{t,k}
    \end{equation}
    We are given the initial condition $\Delta_\tau > 0$.
    We analyze the evolution of this gap for $t > \tau$ conditioned on the event that $i$ retains rank 1.

    \begin{enumerate}
        \item \textbf{Drift of the relevant document:} Since $i$ holds rank 1, the GP-\textsc{Belief} policy assigns it to the position with maximal diagnosticity.
        The expected increment for $i$ is strictly positive:
        \begin{equation}
            \mathbb{E}[\lambda_{t,i} - \lambda_{t-1,i}] = \mu_{(1)} > 0
        \end{equation}
        
        \item \textbf{Drift of distractors:} Any distractor $k \in \mathcal{K}$ is assigned to a position $j = \sigma_t(k)$.
        Since $Z_k=0$, the expected update is the negative KL divergence between the likelihoods (negative drift):
        \begin{align*}
            \mathbb{E}[\lambda_{t,k} - \lambda_{t-1,k}]
             &= -D_{KL}(P_{Z=0, j} || P_{Z=1, j}) \\
             &\leq 0            
        \end{align*}
    \end{enumerate}
    
    Thus, the gap $\Delta_t$ evolves as a random walk with a strictly positive expected drift (since the relevant document accumulates positive signal while distractors accumulate negative or neutral signal). 
    
    From the theory of random walks with positive drift, starting at $\Delta_\tau > 0$, there exists a strictly positive probability that the walk never crosses zero (i.e., $\Delta_t > 0$ for all $t > \tau$).
    On this event, document $i$ never loses the top rank.
    Consequently, it remains permanently assigned to the optimal position, maintaining the maximal information rate $\mathcal{R}_{\textsc{GP-Belief}}(t) = \mu_{(1)}$ for all subsequent rounds.
\end{proof}

\section{Calibration}
\label{app:calibration}

\begin{figure*}
    \centering
    \includegraphics[width=\linewidth]{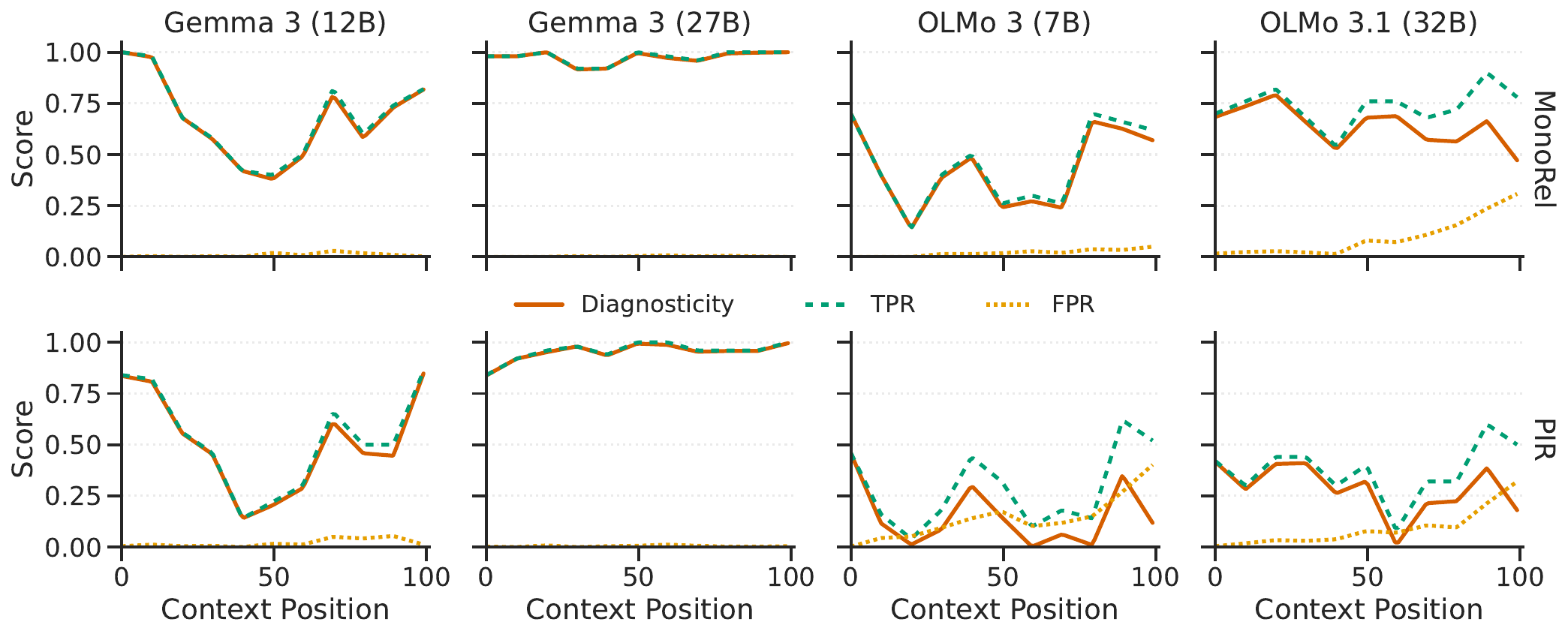}
    \caption{Calibrated position profiles across models and tasks ($N=100$). 
    Each panel shows TPR (solid), FPR (dashed), and diagnosticity $|TPR - FPR|$ (dotted) by context position.
    Gemma-3-12B and OLMo models exhibit pronounced primacy and recency effects, while Gemma-3-27B shows a flat profile with minimal position bias, explaining why strategic reordering offers no advantage for this model.
    }
    \label{fig:calibration_facet_N100}
\end{figure*}

\GP{} requires position-wise detector parameters $(\mathrm{TPR}_j, \mathrm{FPR}_j)$ estimated once per model.
Here we describe the calibration procedure and validate that simple fixed-grid sampling suffices.

\textbf{Calibration procedure.}
For each dataset (MonoRel, PIR), we construct calibration instances to estimate position-wise detector parameters.
Each instance is formed by selecting one sample as the known-relevant ``gold'' document and drawing $N-1$ distractors from other samples in the same dataset.
We place the gold at a target position $j$, fill remaining positions with distractors, and query the model once, yielding a set of cited document IDs.
This induces binary observations for every document--position pair: a true positive when the gold is cited, a false positive when a distractor is cited, and analogous negatives otherwise.

For each position, we repeat this procedure $50$ times with different gold documents and distractor sets.
We average over $10$ independent trials to obtain stable estimates of $\mathrm{TPR}_j$ and $\mathrm{FPR}_j$.
Diagnosticity $d_{\mathrm{diag}}(j) = |\mathrm{TPR}_j - \mathrm{FPR}_j|$ is then computed from these estimates.

\textbf{Conditional independence assumption.}
We adopt a conditional-independence approximation: given the true labels $\mathbf{Z}$ and the assignment $\sigma_t$, citation outcomes are independent across documents within a round.
In practice, citation events may exhibit within-round dependencies due to limited citation budgets or answer structure constraints.
We therefore interpret $(\mathrm{TPR}_j, \mathrm{FPR}_j)$ as position-wise \emph{marginal} detection rates under a fixed protocol.
Under this interpretation, within-round dependencies are absorbed into the calibrated parameters rather than modeled explicitly.
This approximation is conservative: if citations are negatively correlated (e.g., due to a citation budget), our independence assumption overestimates false positive rates, leading to more cautious belief updates.

\section{Asymmetric Document-Position Cases}
\label{app:asymmetric}

The main text presents the balanced case where the number of documents $N$ equals the number of context positions.
In practice, these quantities may differ: retrieval may return more documents than fit in context, or the context window may have unused capacity.
We describe how to reduce asymmetric cases to the balanced setting via dummy items.

\textbf{More documents than positions ($N_{\text{doc}} > N_{\text{pos}}$).}
When the retrieved set exceeds context capacity, we cannot place all documents in a single query.
We introduce $N_{\text{doc}} - N_{\text{pos}}$ \emph{dummy positions} with parameters $\mathrm{TPR} = \mathrm{FPR} = 0$.
Documents assigned to dummy positions are excluded from the current query and receive no observation, leaving their beliefs unchanged.
The greedy matching then operates over $N_{\text{doc}}$ documents and $N_{\text{doc}}$ positions (real + dummy).
Under \textsc{GP-Belief}, the lowest-belief documents are assigned to dummy positions and effectively ``wait'' until higher-belief documents are resolved.
Under \textsc{GP-Entropy}, the lowest-uncertainty documents wait instead.

This reduction preserves the algorithm's convergence guarantee as long as every document is eventually placed at an informative real position.
In practice, with $T$ rounds and $N_{\text{pos}}$ real positions per round, we can test up to $T \cdot N_{\text{pos}}$ document-position pairs, which suffices for convergence when $T$ is moderate.

\textbf{More positions than documents ($N_{\text{pos}} > N_{\text{doc}}$).}
When context capacity exceeds the document set, some positions remain unfilled.
We introduce $N_{\text{pos}} - N_{\text{doc}}$ \emph{dummy documents} with fixed belief $b = 0$ (known irrelevant).
These are assigned to the lowest-diagnosticity positions, ensuring that real documents occupy the most informative slots.
Dummy documents do not participate in belief updates; their presence simply pads the assignment to achieve a balanced matching.

\textbf{Implementation note.}
In our experiments, we use $N = N_{\text{doc}} = N_{\text{pos}} = 100$ throughout, so these reductions are not exercised.
However, the framework naturally extends to production settings where retrieval set sizes vary across queries.
The key insight is that dummy items with degenerate parameters ($\mathrm{TPR} = \mathrm{FPR}$ for positions, $b = 0$ or $b = 1$ for documents) do not affect the information-theoretic properties of the algorithm.

\section{Profile Stability Across Tasks}
\label{app:profile_stability}

\S\ref{subsec:Transferability} demonstrated that \GP{} remains effective under cross-calibration, suggesting that diagnosticity profiles are structural properties of the model rather than task-specific artifacts.
Here we quantify this stability directly by measuring the rank correlation between profiles calibrated on MonoRel versus PIR.

\begin{table}[ht]
    \centering
    \small
    \caption{
    Spearman's $\rho$ and Kendall's $\tau$ between diagnosticity profiles calibrated on MonoRel vs.\ PIR.
    \textcolor{red}{Red} entries are not statistically significant ($p > 0.05$); these correspond to degenerate cases where the model exhibits minimal position bias.
    }
    \begin{tabular}{l c c}
        \toprule
        Model & $\rho$ & $\tau$ \\
        \midrule
        \multirow{1}{*}{OLMo-3-7B-Think}
            & 0.595 & 0.455 \\
        \midrule
        \multirow{1}{*}{OLMo-3.1-32B-Think}
            & 0.488 & 0.315 \\
        \midrule
        \multirow{1}{*}{Gemma-3-12B-IT}
            & 0.959 & 0.842 \\
        \midrule
        \multirow{1}{*}{Gemma-3-27B-IT}
            & \textcolor{red}{0.102} & \textcolor{red}{0.112} \\
        \bottomrule
    \end{tabular}
    \label{tab:profile_correlation}
\end{table}

Tab.~\ref{tab:profile_correlation} reports Spearman's $\rho$ and Kendall's $\tau$ for each model at $N=100$ positions.
Models with exploitable position bias show strong rank correlations ($\rho > 0.49$), indicating that the relative ordering of high- vs.\ low-diagnosticity positions is preserved across tasks. 
This stability explains why cross-calibration incurs only modest performance degradation 
(Fig.~\ref{fig:transfer}): the algorithm primarily relies on \emph{which} positions are 
reliable, not their precise TPR/FPR values.



\end{document}